\documentclass{article}

\usepackage[numbers,compress,sort]{natbib}

\usepackage[dblblindworkshop, final]{neurips_2026}

\usepackage{float}

\usepackage[table]{xcolor}
\definecolor{bestbg}{gray}{0.82}
\definecolor{secondbg}{gray}{0.93}

\usepackage[utf8]{inputenc}
\usepackage[T1]{fontenc}
\usepackage{hyperref}
\usepackage{url}
\usepackage{booktabs}
\usepackage{amsfonts}
\usepackage{amsmath}
\usepackage{nicefrac}
\usepackage{microtype}
\usepackage{xcolor}
\usepackage{graphicx}
\usepackage{multirow}
\usepackage{lipsum}
\usepackage{comment}

\workshoptitle{ML4Molecules: Agentic Systems for Molecular Sciences}

\title{LLMs Depend on Specialist Tools for Out-of-Distribution Generalization in MS/MS Structural Elucidation}

\author{
  {Gabriel Asher} \quad
  {Cailum M. K. Stienstra} \quad
  {Tomo Oga} \\
  \textbf{Devesh Shah} \quad
 \textbf{Antonio H. de O. Fonseca} \quad
  \textbf{Niall O'Connor} \quad
  \textbf{Michael Widrich}\\
  Matterworks, Inc. \\ Somerville, MA 02143, USA \\
  \texttt{gabriel@matterworks.ai}
}

\begin{document}
\maketitle

\begin{abstract}
A frontier language model was recently reported to solve 1D NMR structural elucidation without tools. We ask whether this capability transfers to tandem mass spectrometry (MS/MS) in the \textit{de novo} setting, where the model must propose a structure directly from the spectrum without a candidate list or reference library. To separate reasoning from recall, we build a 100-analyte benchmark from MassSpecGym, split by each analyte's \textit{textual footprint} into in-distribution (well documented in literature) and out-of-distribution (essentially undocumented) arms. We hold the model, harness, and output format fixed and vary only the LLM's access to domain-specific tools. Without tools, a frontier LLM reaches 52\% Top-1 accuracy in-distribution but only 2\% out-of-distribution. Only access to \textit{de novo} structure generators improves out-of-distribution accuracy, raising it to as much as 38\% Top-1, and the resulting agent outperforms the standalone \textit{de novo} models both in and out of distribution. These results suggest that LLMs are not effective at \textit{de novo} MS/MS structural elucidation on their own, and that further gains depend on the specialist tools available to the agent rather than on the LLM itself.\end{abstract}

\section{Introduction}\label{sec:intro}
Tandem mass spectrometry (MS/MS) is the dominant approach for identifying unknown small molecules in complex mixtures. Its sensitivity extends to trace quantities that are inaccessible to nuclear magnetic resonance (NMR) spectroscopy \citep{Heiles2021AdvancedMS, Strynar2015PFECAs}. MS/MS takes an ionized \textit{precursor} compound and breaks it into fragments, each of which contains a mass and an intensity. Inferring a compound's identity from its spectrum, a process called \textit{structural elucidation}, is a central problem in the field of mass spectrometry. This task remains largely unsolved, with most spectra detected in a sample never assigned to a structure. These spectra are known colloquially as the 
``dark matter'' of metabolomics \citep{qiang2026anticipation}.

The standard method for structural elucidation is \textit{library matching}, which compares a query spectrum against a library of experimentally measured reference spectra and ranks the candidate identities by fragmentation-pattern similarity \citep{Kind2018AccurateMassMSMS}. The coverage of library matching is bounded by the size of its spectral libraries, which typically contain $10^{4}$--$10^{5}$ distinct structures \citep{massspecgym2024}, orders of magnitude fewer than the ${\sim}10^{8}$ compounds cataloged in PubChem \citep{kim2025pubchem, COLEY2021133}. Furthermore, library-based approaches cannot identify a compound that has never been measured, a category that includes many novel drug metabolites and environmental transformation products. Identifying these compounds requires a class of structural elucidation named \textit{de novo generation}. \textit{De novo} elucidation, a form of \textit{inverse structural elucidation}, generates a candidate structure directly from an MS/MS spectrum, and is the most challenging formulation of structural elucidation. On MassSpecGym, the community-accepted benchmark for this task,
the state-of-the-art \textit{de novo} Top-1 accuracy is 23.9\% \citep{bohde2026frigidscalingdiffusionbasedmolecular, zhao2026ms}.

Recent work reports strong zero-shot performance of frontier LLMs on the inverse elucidation problem in NMR spectroscopy. In \textit{Making Claude a Chemist} \citep{kamber2026claudechemist},
Claude Opus~4.7 was given fifteen 1D inverse elucidation problems,
each supplying $^{1}$H and $^{13}$C peak lists, molecular formulae,
and, for the seven hardest spectra, the reaction's starting material. With no tools and no chemistry-specific fine-tuning, a setting that mirrors \textit{de novo} elucidation, it returned the published structure on at least two of three attempts for every compound.
Because the compounds were taken from preprints posted after the model's training cutoff,
the result cannot be attributed to memorization.

This work asks whether the zero-shot structural elucidation performance reported for Claude Opus on NMR \citep{kamber2026claudechemist} translates to MS/MS. This translation is not guaranteed, as chemical shift in NMR reflects each nucleus's local environment and is approximately additive, making the substructure-to-shift mapping compositional \citep{Stienstra2025NMRNetworking}. By contrast, MS/MS fragmentation is stochastic and non-local, proceeding through charge migration, rearrangement, and competing dissociation channels \citep{Heiles2021AdvancedMS}, so fragments vary with adduct and instrument \citep{wang2025neural, massspecgym2024}. Unlike NMR shifts, MS/MS spectra therefore cannot be reliably predicted, or inverted, from substructure rules alone.

We additionally explore whether an LLM with tools, an agent, can improve \textit{de novo} elucidation performance. In other chemistry domains, tools cover capabilities that a base model lacks \citep{bran2024chemcrow, boiko2023coscientist}, but the benefit is task-dependent and can be negative for competent LLMs \citep{yu2024chemtoolagent}. No equivalent attribution exists for MS/MS, where published comparisons of agents with and without tools reduce to a single all-or-nothing contrast between the full toolbox and none of it \citep{wang2026gnps2agentic}. Furthermore, without a careful ablation of the tools used in an agent, it is challenging to determine whether an agent came to an answer via reasoning over its tool outputs, or memorization (recall) over its training data.

\textbf{Contributions.}
To evaluate the \textit{de novo} MS/MS elucidation capabilities of agentic LLMs, we contribute the following.
(a) We construct a 100-spectrum MS/MS benchmark from the MassSpecGym test fold. This benchmark contains two sets of spectra matched on acquisition conditions but with a different literature footprint,
separating memorization from reasoning-based elucidation.
(b) We show that, without tools, the LLM's accuracy tracks a compound's literature footprint rather than the information in its spectrum. It reaches 52\% Top-1 accuracy on well-documented in-distribution (ID) compounds but only 2\% on out-of-distribution (OOD) compounds.
(c) We attribute performance to individual tools by ablation under a fixed harness,
finding that subformula annotation and forward prediction only improve performance on in-distribution analytes,
while a \textit{de novo} generator raises out-of-distribution accuracy to 12--38\%.
 (d) We show that this out-of-distribution gain is bounded by the candidates the generator proposes, with the LLM acting as a better re-ranker.

\section{A benchmark split by textual footprint}
\label{sec:benchmark}

\begin{figure}
    \centering
    \includegraphics[width=0.9\linewidth]{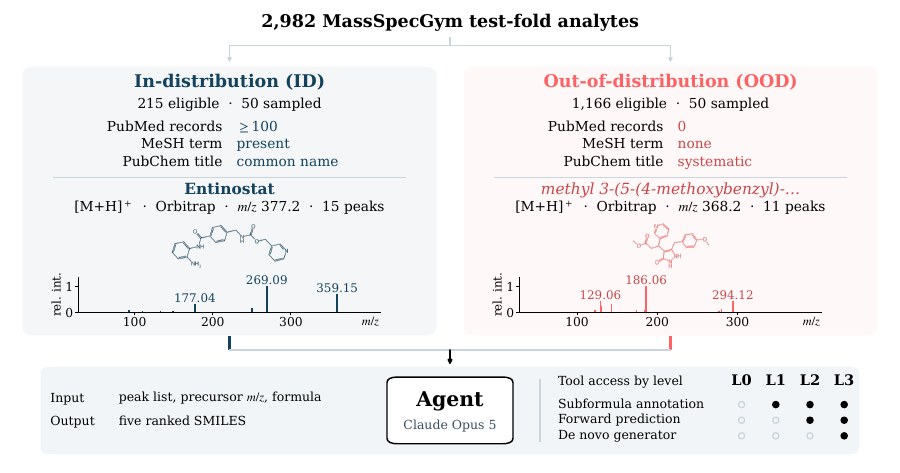}
    \caption{Method overview. Analytes are split by documentation and matched 1:1 on acquisition; one member of a matched pair is shown per benchmark arm with its structure and spectrum. All experiments share one agent, prompt, and scorer, differing only in tool access.}
    \label{fig:method}
\end{figure}

\textbf{Data.}\label{sec:data-methods}
All spectra were selected from the test fold (17{,}556 spectra) of MassSpecGym v1.5~\citep{liu2026massspecgym},
since the tools given to our agent were trained on this benchmark's predefined split.
Structures are standardized with the pipeline of \citet{gupta2026comprehensive},
which strips stereochemistry, neutralizes charges, and canonicalizes tautomers,
yielding 2{,}992 unique test-fold analytes (Appendix~\ref{app:textual_splitting}).
Unique molecular identity is defined by the 14-character InChIKey connectivity block, because MS/MS alone cannot resolve stereoisomers.

\textbf{Benchmark Creation.}
This work seeks to compare in-distribution versus out-of-distribution performance of LLMs on structural elucidation. However, frontier models rarely disclose their training data, making it difficult to determine whether they have encountered a given compound. We therefore use a molecule's \emph{textual footprint}, defined as the number of PubMed records linked to the compound in PubChem~\citep{kim2016literature}, as a proxy for whether an analyte exists in an LLM's training data. The intuition behind this choice is that frontier models are extensively trained on scientific literature, and compounds that are more prevalent in literature are more likely to exist in an LLM's training dataset, thus becoming in-distribution. Our benchmark is split into two arms (Figure~\ref{fig:method}, Appendix Table~\ref{tab:curation}). The \emph{textually out-of-distribution} (OOD) arm selects analytes with zero linked PubMed records, no Medical Subject Headings (MeSH) entry, and a systematic (IUPAC) PubChem title, whereas the \emph{textually in-distribution} (ID) arm selects analytes with at least 100 linked PubMed records, a MeSH entry, and a common-name title. Note that, given our benchmark setup, ``OOD'' here refers to the textual distribution alone and does not imply chemical novelty.

In order to isolate training distribution as our independent variable, we control for ``spectral difficulty'' in selecting our benchmark analytes. We randomly select 50 OOD analytes from our candidate set and pair them to the closest unused ID analyte by precursor $m/z$, fragment-peak count, adduct, and instrument.
Each analyte contributes one spectrum, so in this text we use ``analyte'' and ``spectrum'' interchangeably. The 100 analytes were fixed before any experiment was run (see Appendix~\ref{app:textual_splitting} for details).

\textbf{Task and Metrics.}
All experiments use the same task. Given the (\textit{m/z}, \textit{intensity}) peaks of an MS/MS spectrum, the precursor \textit{m/z}, and the precursor's neutral molecular formula, identify the precursor molecule, returning $k=5$ ranked candidate structures as SMILES. We evaluate performance using Top-1 and Top-5 accuracy, where a prediction is counted as correct if the 2D InChIKey of any of the Top-$k$ generated molecules matches that of the ground-truth precursor. The 2D InChIKey is the first 14-character block of the full InChIKey, and encodes molecular connectivity but not stereochemistry. We additionally measure molecular similarity of generated molecules against their ground truth via Tanimoto@1 over 2048-bit ECFP4 fingerprints and MCES@1, a graph edit distance.

\textbf{Harness.}
In order to adequately assess tool contribution to structural elucidation, the agent harness is fixed for all experiments and only the tool set varies. Query spectra, model, output contract, parser, scorer, sampling parameters, and turn budget are identical across experiments.
All experiments run Claude Opus~5 at high reasoning effort with a
64{,}000-token output allowance and a ceiling of twenty tool calls per query.
Every tool run shares one system prompt and differs only in the attached tool definitions,
so differences among them are attributable to the tools alone.
The LLM baseline experiment uses a modified prompt which omits tool descriptions. Each analyte is elucidated with fresh context, and all tool outputs are screened to ensure no ground-truth annotations were leaked to the agent.
Analytes are re-run under the same configuration until every experiment reports 100 usable answers (Appendix~\ref{app:harness}). 

\textbf{Tools.}
Three tools are available to the agent:
\emph{subformula annotation} returns a subformula for every peak in an MS/MS spectrum that is consistent with the parent formula,
\emph{forward prediction} simulates an MS/MS spectrum from a SMILES string,
and \emph{de novo generation} attempts to identify structures directly from the spectrum, returning a list of candidate SMILES. Subformula annotation and forward prediction let an agent test candidate identities. \textit{De novo} generators instead propose those identities, shifting the agent's work from generating candidates to evaluating them. These tools are provided to the agent via two interchangeable tool suites. The \emph{public} suite uses MIST-CF's subformula annotation script \citep{goldman2024mist} for subformula annotation, GLACIER \citep{wang2026glacier} for forward prediction, and MS-GPT \citep{zhao2026ms} and FRIGID \citep{bohde2026frigidscalingdiffusionbasedmolecular} for \textit{de novo} generation. As of the time of experimentation (August 2026), MS-GPT and FRIGID are the current and previous state-of-the-art \textit{de novo} identification models on the MassSpecGym benchmark. The \emph{in-house} suite substitutes proprietary models for subformula annotation and \textit{de novo} generation while retaining GLACIER. We denote our in-house generator \textsc{LSM-Gen}, a \textit{de novo} model from our LSM (Large Spectral Model) family \citep{asher2025lsm}, retrained for this paper. 

All three generators use spectrum encoders trained on the same spectra (FRIGID and MS-GPT share a single MIST checkpoint; Appendix~\ref{app:mist_leakage}), but their decoders are trained on different molecular datasets. FRIGID and MS-GPT pretrain their decoders on 100M--230M structures from the SAFE-GPT molecule collection \citep{noutahi2024gotta}, excluding test structures only by exact 2D InChIKey match \citep{liu2026massspecgym}. \textsc{LSM-Gen} instead uses a stricter split, the smaller 50M-structure leakage-reduced dataset of \citet{liu2026massspecgym}, which also removes any molecule with ECFP4 Tanimoto similarity $\geq 0.7$ to a test structure (Appendix~\ref{app:decoder_decontamination}).

\textbf{Experiments.} 
Experiments are cumulative, each adding one tool at a time. \textbf{L0} is the LLM-only baseline. Given that no tools are used for this experiment, whatever performance it achieves comes only from pattern recognition, chain-of-thought reasoning, and memorization. \textbf{L1} adds subformula annotation, removing peak assignment from the model's own arithmetic. \textbf{L2} adds forward prediction, used to test candidate structures once they are predicted. \textbf{L3} adds \textit{de novo} generation, which provides the agent with a set of candidate structures that it can then test. At L3 the generated structures are added to the agent's candidate set, not swapped in for it. Each generator model is also run standalone as a non-agentic baseline.

\section{Results}
\label{sec:results}

\newcommand{\sig}[1]{\textsuperscript{\normalfont #1}}
\newcommand{\sigw}[1]{\textsuperscript{\normalfont (#1)}}
\newcommand{\nodisc}{\textsuperscript{\normalfont\dag}}
\begin{table}[t]
  \caption{The ablation ladder on $N{=}100$ benchmark analytes (ID and OOD, $n{=}50$ each).}
  \label{tab:main}
  \centering
  \scriptsize
  \setlength{\tabcolsep}{3pt}
  \begin{tabular}{ll ccc ccc ccc ccc}
    \toprule
    & & \multicolumn{3}{c}{Top-1 $\uparrow$} & \multicolumn{3}{c}{Top-5 $\uparrow$}
      & \multicolumn{3}{c}{Tanimoto@1 $\uparrow$} & \multicolumn{3}{c}{MCES@1 $\downarrow$} \\
    \cmidrule(lr){3-5}\cmidrule(lr){6-8}\cmidrule(lr){9-11}\cmidrule(lr){12-14}
     & Model & ID & OOD & All & ID & OOD & All & ID & OOD & All
      & ID & OOD & All \\
    \midrule
    \multicolumn{14}{l}{\textit{Non-agentic reference (formula given)}} \\
    -- & Public de novo (FRIGID) & 24\sigw{*} & 6 & 15 & 30\sigw{*} & 12 & 21 & 0.51 & 0.38 & 0.44 & 9.4\sigw{*} & 11.6 & 10.5 \\
    -- & Public de novo (MS-GPT) & 48 & 18\sig{*} & 33 & 54 & 30\sig{***} & 42\sig{*} & 0.69 & 0.50\sig{***} & 0.60\sig{**} & 6.2 & 9.8\sig{***} & 8.0 \\
    -- & In-house de novo (\textsc{LSM-Gen}) & \textbf{60} & \textbf{32}\sig{***} & \textbf{46}\sig{**} & \textbf{64} & \textbf{36}\sig{***} & \textbf{50}\sig{**} & \textbf{0.73} & \textbf{0.58}\sig{***} & \textbf{0.66}\sig{***} & \textbf{4.9} & \textbf{8.0}\sig{***} & \textbf{6.5}\sig{**} \\
    \midrule
    \multicolumn{14}{l}{\textit{Agentic ladder (formula given)}} \\
    L0 & LLM baseline & 52 & 2 & 27 & 54 & 2 & 28 & 0.66 & 0.31 & 0.48 & 6.0 & 13.3 & 9.6 \\
    L1 & $+$ subformula annotation & 62 & 2 & 32 & 64 & 2 & 33 & 0.74 & 0.28 & 0.51 & 4.6 & 13.6 & 9.1 \\
    L2 & $+$ forward prediction & 56 & 2\nodisc & 29 & 62 & 2\nodisc & 32 & 0.69 & 0.30 & 0.49 & 5.5 & 12.9 & 9.2 \\
    L3 & $+$ FRIGID & 68 & 12 & 40\sig{*} & 70 & 20\sig{*} & 45\sig{***} & 0.79\sig{*} & 0.45\sig{**} & 0.62\sig{***} & 3.9 & 10.0\sig{***} & 7.0\sig{***} \\
    L3 & $+$ MS-GPT & \textbf{86}\sig{***} & 26\sig{***} & 56\sig{***} & \textbf{90}\sig{***} & 34\sig{***} & \textbf{62}\sig{***} & \textbf{0.90}\sig{***} & 0.51\sig{***} & 0.70\sig{***} & \textbf{1.9}\sig{***} & 9.0\sig{***} & \textbf{5.5}\sig{***} \\
    L3 & $+$ \textsc{LSM-Gen} (in-house tools) & 80\sig{**} & \textbf{38}\sig{***} & \textbf{59}\sig{***} & 82\sig{**} & \textbf{42}\sig{***} & \textbf{62}\sig{***} & 0.84\sig{*} & \textbf{0.57}\sig{***} & \textbf{0.71}\sig{***} & 2.9\sig{*} & \textbf{8.0}\sig{***} & \textbf{5.5}\sig{***} \\
    \bottomrule
  \end{tabular}
  \vspace{3pt}
  \begin{minipage}{\linewidth}
  \scriptsize\raggedright
  Accuracies are percentages. Bold marks the best value per column within each block.
  All rows are tested against L0: McNemar for Top-1 and Top-5, Wilcoxon signed-rank for
  Tanimoto@1 and MCES@1. *~$p<0.05$, **~$p<0.01$, ***~$p<0.001$, unadjusted; stars in
  parentheses indicate the row is significantly \emph{worse} than L0. \dag~McNemar
  undefined: no discordant pairs.
  \end{minipage}
\end{table}

\textbf{Unaided LLMs are retrieval engines.}
At L0 the LLM's Top-1 accuracy is 52\% (26/50) for ID analytes and 2\% (1/50) for OOD analytes (Table~\ref{tab:main}).
Because the benchmark arms do not separate on adduct, precursor $m/z$, or peak count (Appendix~\ref{app:textual_splitting}),
a 50-point gap cannot be attributed to spectral difficulty.
Tanimoto@1 further separates them by a large margin ($0.66$ against $0.31$);
the OOD predictions are not merely mis-ranked but structurally wrong.
In contrast to the LLM baseline's 2\%, the best \textit{de novo} models achieve 18\% and 32\% Top-1 accuracy on the OOD arm. Undocumented compounds are therefore identifiable from their spectra alone, so the LLM's ID/OOD gap is not explained by OOD spectra being less informative. Instead, it suggests that the LLM's in-distribution accuracy reflects memorized knowledge of documented compounds rather than reasoning over the spectrum from first principles.

\textbf{Tools help and \textit{de novo} generators enable OOD.} Subformula annotation (L1) and forward prediction (L2) improve ID Top-1 accuracy from 52\% at L0 to 62\% and 56\% respectively, but leave OOD unchanged at 2\%. Adding a \textit{de novo} generator raises OOD Top-1 to 12\% with FRIGID, 26\% with MS-GPT, and 38\% with \textsc{LSM-Gen}, lifting performance across all 100 analytes to 40\%, 56\%, and 59\% respectively. The two hypothesis-testing tools, forward prediction and subformula annotation, contribute only toward identifying ID analytes, whereas the tools that propose candidate structures enable OOD hits. Forward prediction is also mildly counterproductive where it does act, costing six points of ID Top-1 in L2 relative to L1. Across all experiments, at most 14\% of errors occur in cases where the agent had the correct structure but failed to return it first. In nearly all of the remainder, the structure was never proposed by the agent or its tools (Appendix Figure~\ref{fig:failure-cascade}). This finding suggests that the generation of plausible chemical structures, rather than the ranking of candidates once proposed, is the primary bottleneck of agentic reasoning in \textit{de novo} MS/MS elucidation.

\begin{figure}[t]
    \centering
    \includegraphics[width=0.9\linewidth]{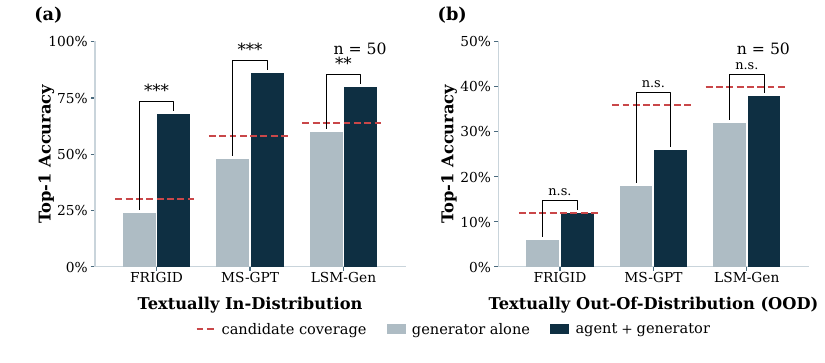}
     \caption{Agent accuracy against its generator's candidate coverage. Top-1 accuracy
    on the \textbf{(a)} ID and \textbf{(b)} OOD splits ($n{=}50$ each). Dashed: coverage, the
    share of spectra whose true structure is anywhere in the generator's pool. Brackets: exact
    McNemar, agent vs.\ generator, unadjusted.}
    \label{fig:coverage}
    \end{figure}

\paragraph{OOD gain is bounded by what the generator proposes.}
The \textit{de novo} models in L3 generate a set of candidates for the agent to interrogate.
The agent's task therefore shifts from generating candidates to ranking them. The agent is good at
candidate re-ranking in both benchmark arms. When the true structure is present in the \textit{de novo} model's candidate set, it correctly ranks this structure as its top answer for 93--97\% of ID analytes and 67--95\% of
textually OOD analytes (Figure~\ref{fig:coverage}). Given that OOD analytes carry no PubMed-linked records or MeSH entry, this re-ranking is unlikely to occur due to recall, and instead demonstrates the agent's reasoning capabilities over its tool outputs. 

The agent may also return structures its generator never produced, so its accuracy can in principle exceed the \textit{generator's candidate coverage}, the fraction of spectra whose true structure appears anywhere in the \textit{de novo} model's generated candidates. On the ID arm it does: 68\% against a coverage of 30\% for FRIGID, 86\% against 58\% for MS-GPT,
and 80\% against 64\% for \textsc{LSM-Gen}. That amounts to 20, 15, and 9 correct structures
the generator tool's pool never held, and each of the three gains over the generator alone is significant
(exact McNemar, $p<0.01$). On the OOD analytes, the agent's own proposed candidates add little. The agent's accuracy with MS-GPT and \textsc{LSM-Gen} no longer clears its generator's coverage, reaching 26\% against 36\% for
MS-GPT and 38\% against 40\% for \textsc{LSM-Gen} (Figure~\ref{fig:coverage}). FRIGID is the one case where the agent does
reach outside the candidate pool, but the two structures it gains are offset by two in-pool structures
it loses, leaving 12\% against a coverage of 12\%. This difficulty generating identity candidates out-of-distribution suggests that an agent is best equipped to adjudicate over the candidate identities provided by its tools, rather than to attempt to generalize on its own.
\section{Discussion and Limitations}
\label{sec:discussion}\label{sec:limitations}

Our work shows that, without tools, a frontier model's MS/MS \textit{de novo} identification accuracy tracks how much has been written about a compound rather than what its spectrum contains. This shortcoming can be addressed by giving an agent a \textit{de novo} generator tool, over whose candidates an agent is an effective re-ranker. The approach presented here is not without limitations, which we outline below.

Re-ranking improves elucidation accuracy on the hardest spectra, those that require \textit{de novo} identification, but at roughly twenty times the cost of the generator alone (Appendix~\ref{app:costs}). Samples may hold thousands of MS/MS spectra, so on its own the agent studied here is prohibitively expensive for untargeted metabolomics workflows. We suggest that a practical agent would instead resolve most spectra by library matching and in-silico retrieval, pass most of the remainder to the generator alone, and reserve the \textit{de novo} elucidation workflow studied here for the few spectra that matter most.

The difference in ID/OOD performance is not unique to the LLM, which limits what these experiments can attribute to it. In the LLM baseline experiment, Top-1 accuracy falls from 52\% on ID analytes to 2\% on OOD, a factor of 26. However, the performance of all three generator-only runs falls as well, though by a much smaller factor of roughly two to four (24\% to 6\% for FRIGID, 48\% to 18\% for MS-GPT, and 60\% to 32\% for \textsc{LSM-Gen}). One explanation is that ID analytes have closer structural analogues in the generators' encoder training data. All three \textit{de novo} models condition on encoders trained with the same PubChem-derived dataset of synthetic spectra, from which test structures are removed by exact InChIKey match only, leaving close analogues in place. Nearest-neighbor similarity of test-set analytes to that dataset splits by benchmark arm (ECFP4 Tanimoto median 0.90 ID against 0.82 OOD, Appendix~\ref{app:mist_leakage}). Future work should therefore remove close analogues from the encoder's training data, not just exact matches. Importantly, this is not direct leakage since the synthetic spectra in the encoder's training dataset are simulated by a forward model trained only on MassSpecGym's training split, so the models never see an experimental spectrum of any test compound.

Our results carry further limitations. Firstly, textual footprint is a proxy for out-of-distribution which we cannot verify without observing the LLM's training datasets. Training datasets of frontier models are undisclosed and these analytes do exist in PubChem, so a compound with no literature record may still have been seen. Secondly, other forms of MS/MS leakage in the LLM may exist. In Appendix~\ref{app:additional_results}, we observe a significant improvement in performance on analytes whose PubChem pages contain MS/MS spectra. Thirdly, we evaluate a single model at a single date, without sweeping providers, temperatures, or token budgets. Finally, Opus~5 is not uniformly the strongest LLM on scientific tasks \citep{koch2026bixbench3}, so the experiment's absolute numbers should be read as one configuration rather than a frontier.

\section{Conclusion}
\label{sec:conclusion}
Unlike in NMR, a frontier model alone cannot perform \textit{de novo} structural elucidation of tandem mass spectra. Opus 5 zero-shot identifies 27\% of our benchmark spectra, but these are almost exclusively well-cited compounds, and Top-1 accuracy falls from 52\% to 2\% between in- and out-of-distribution analytes. When integrated with specialist tools in an agentic framework, an LLM outperforms state-of-the-art \textit{de novo} generator models both in and out of distribution, although its out-of-distribution performance remains limited by the candidates its \textit{de novo} generator provides. Future efforts to annotate a larger share of the dark matter of metabolomics will therefore require both better \textit{de novo} models and more sophisticated agents that draw on a wider range of structural elucidation tools.

\begin{ack}
All authors are employees and shareholders of Matterworks, Inc. 

\end{ack}

\bibliographystyle{plainnat}
\bibliography{References}
\appendix

\section{Related work}
\label{sec:related}

\paragraph{Traditional MS/MS identification relies on library matching.} Routine annotation matches an observed spectrum against reference libraries by cosine, modified-cosine, or spectral-entropy similarity, and is therefore restricted to compounds a library already contains. Structural retrieval relaxes this constraint by matching predicted fingerprints \citep{goldman2024mist, duhrkop2019sirius} or through \textit{in-silico} spectra \citep{wang2021cfm, wang2025neural, wang2026glacier} to structure databases like PubChem. Both approaches assume that the structure of interest already exists in some library or database, which is an assumption that does not hold for novel or uncharacterized analytes.

\paragraph{\textit{De novo} molecular generation.} \textit{De novo} structural elucidation methods overcome the limitations of library matching by generating a structure directly from the spectrum. Early fingerprint-to-structure decoders \citep{stravs2022msnovelist} were followed by formula-conditioned generative models, including diffusion over molecular graphs and sequences \citep{bohde2026frigidscalingdiffusionbasedmolecular}. Current state-of-the-art public models score roughly 24\% Top-1 accuracy on chemically OOD test sets \citep{zhao2026ms}.  In order to read our results against the state-of-the-art, we used the top two \textit{de novo} models available at the time of experimentation in August 2026.

\paragraph{Language models and agents on spectra.}
Off-the-shelf LLMs have previously been observed to fail outright at direct spectra-to-structure prediction: prompted with chain-of-thought on MassSpecGym, four LLMs return no exact match at Top-10 accuracy, against 4.25\% for a spectrum-trained diffusion baseline \citep{wang2026offtheshelf}. Agentic framings have been more productive. The GNPS2 framework couples an LLM to tools for MS/MS alignment, formula decomposition, structural enumeration, substructure mapping, spectral prediction, and validation against mass, adduct, and retention time; it recovered several previously unannotated metabolites, later confirmed experimentally \citep{wang2026gnps2agentic}. MSAgent orchestrates an LLM with more than fifty domain-specific tools, reporting MRR gains above $10\%$ on CASMI retrieval and roughly $40\%$ mean Tanimoto improvement when refining \textit{de novo} outputs on CANOPUS, both against a tool-free GPT-4o arm \citep{li2026msagent}. MS4MS pairs a fine-tuned domain model with a formula agent and an identification agent that cosine-matches spectral libraries and reasons over the retrieved neighbors, reporting $92.0\%$ Top-1 formula accuracy on the same MassSpecGym test fold used here \citep{guo2025ms4ms}. MetaboT routes an LLM over a metabolomics knowledge graph through entity resolution, schema-aware SPARQL generation, and query refinement, reporting $83.7\%$ accuracy on forty-nine expert-authored questions against $8.2\%$ for single-shot GPT-4o \citep{bekbergenova2026metabot}. MetaboAnnotate uses a multi-agent setup which predicts molecule identities using consensus voting over three orthogonal forward prediction (in-silico fragment generation) tools, reporting $36.5\%$ and $28.2\%$ Top-1 retrieval on CASMI 2016 and 2022 respectively \citep{zhou2026metaboannotate}.  Three limits follow from these existing agents. First, each agent couples retrieval or database lookup to whatever generative capability it has and reports a single identification number over the combination, so none isolates \textit{de novo} generation performance. Second, each is evaluated in a tool-maximal configuration, compared either against an external baseline or against the same model stripped of every tool at once. Therefore, the marginal contribution of any one component cannot be measured. Third, every tool-free arm in this literature, including the chain-of-thought benchmark above, runs on GPT-4o, GPT-5 nano, Claude-3.5-Sonnet, or a fine-tuned 14B model, so the systems are not measured using a frontier model.

\paragraph{Tool attribution and memorization.}
The effect of domain-specific tools has been measured for more general chemistry agents. ChemCrow
and Coscientist show that tool augmentation can produce capabilities the base model lacks
\citep{bran2024chemcrow, boiko2023coscientist}, while ChemToolAgent finds the effect strongly
task-dependent: specialized tasks benefit substantially, whereas tools add little or even
hurt on tasks the base LLM already handles \citep{yu2024chemtoolagent}. 

Memorization in LLMs has also been studied directly. The likelihood that a model memorizes a fact grows log-linearly with how often that fact is duplicated in pretraining \citep{carlini2023quantifying}, and accuracy on factual questions tracks how many pretraining documents mention both the question and its answer, a dependence that scaling alone is unlikely to remove \citep{kandpal2023longtail}. Retrieval augmentation helps most on questions about less popular subjects \citep{mallen2023whennot}. Our agent experiments test this same interaction in a setting where the retrieved object is a molecular structure rather than a fact.

\paragraph{Textual footprint for structural elucidation.}
Note that using PubMed citation counts in MS/MS identification has precedent: MetFrag uses PubMed and patent counts to prioritize well-documented candidates~\citep{ruttkies2016metfrag}. Unlike MetFrag, however, we use documentation not as a ranking signal but as a proxy for how likely a model is to recall a compound rather than identify it from spectral evidence.

\section{Textual Splitting Procedure}
\label{app:textual_splitting}

\begin{table}[t]
  \caption{Benchmark curation. }
  \label{tab:curation}
  \centering
  \small
  \begin{tabular}{lrr}
    \toprule
    & Removed & Remaining \\
    \midrule
    \multicolumn{3}{l}{\textit{Spectra}} \\
    MassSpecGym v1.5 test fold & & 17{,}556 \\
    \quad absent from the candidate pool & 107 & 17{,}449 \\
    \quad fewer than four fragment peaks & 1{,}900 & 15{,}549 \\
    \addlinespace
    \multicolumn{3}{l}{\textit{Analytes}} \\
    Standardized test-fold analytes & & 2{,}992 \\
    \quad unresolved against PubChem & 10 & 2{,}982 \\
    \addlinespace
    \multicolumn{3}{l}{\textit{Out-of-distribution arm}} \\
    \quad zero linked PubMed records & 1{,}692 & 1{,}290 \\
    \quad and no MeSH annotation & 2 & 1{,}288 \\
    \quad and a systematic (IUPAC-style) title & 112 & 1{,}176 \\
    \quad and an eligible spectrum retained & 10 & 1{,}166 \\
    \addlinespace
    \multicolumn{3}{l}{\textit{In-distribution arm}} \\
    \quad at least 100 linked PubMed records & 2{,}714 & 268 \\
    \quad and a MeSH annotation & 31 & 237 \\
    \quad and a common-name title & 17 & 220 \\
    \quad and an eligible spectrum retained & 5 & 215 \\
    \addlinespace
    \multicolumn{3}{l}{\textit{Benchmark}} \\
    1:1 matched on adduct, instrument, mass, and peaks & & 50 pairs \\
    One median-peak spectrum per analyte & & 100 \\
    \bottomrule
  \end{tabular}
\end{table}

This work interrogates LLM performance on MS/MS structural elucidation across in- and out-of-distribution settings. In order to estimate whether an analyte is in-distribution to a language model's training dataset, we measure its \textit{textual footprint}. To measure this footprint, we construct a local snapshot of PubChem using the bulk NCBI distributions for InChIKeys, synonyms, titles, PubMed associations, MeSH terms, patents, and depositor records. Analytes from MassSpecGym are then joined to these PubChem entries via their 2D InChIKey connectivity block (first 14 characters of the InChIKey). 10 out of 2{,}992 MassSpecGym test-fold analytes cannot be joined because their 2D InChIKey after analyte standardization does not exist in PubChem. Multiple PubChem CIDs may map to one 2D InChIKey, so we measure literature prevalence using the maximum number of linked PubMed records among CIDs sharing the same 2D InChIKey block.

We then construct a 100-analyte benchmark containing two sets of 50 analytes (Table \ref{tab:curation}). The \textit{textually OOD} arm contains compounds with zero linked PubMed records, no MeSH annotation, and a PubChem display title classified as systematic (IUPAC-like), whereas the \textit{in-distribution} arm requires at least 100 linked PubMed records, a MeSH annotation, and a common-name title. MassSpecGym often holds several spectra for the same analyte, so we only keep the spectrum with the median peak count to ensure (analyte, spectrum) uniqueness. Importantly, ``OOD'' here refers specifically to the \textit{textual distribution}: these compounds lack the literature indexing and common names that could give a language model an obvious memorized structure--name association. It does not imply that they are chemically out of distribution for a molecular structure model, or that they are absent from the LLM's training data altogether. 

Spectra with three or fewer fragment peaks are removed before sampling. They carry too
little structural information to be reliably interpreted. What remains is
1{,}166 eligible out-of-distribution analytes and 215 in-distribution ones. In order to keep \textit{textual distribution} as our primary independent variable, we control for how we sample spectra between ID and OOD datasets. We sample 50 out-of-distribution analytes under a fixed random seed and pair each with the closest in-distribution analyte by precursor $m/z$ and fragment-peak count, and where possible by adduct (matched in 48 of 50 pairs) and instrument type (41 of 50). Across the 100 spectra, the two arms do not differ in adduct ($p{=}0.68$), precursor $m/z$ ($p{=}0.98$), or fragment-peak count ($p{=}0.70$). Collision energy was not matched and differs slightly between arms (median 45~eV ID against 30~eV OOD, $p{=}0.045$; recorded for 79 of 100 spectra).
\clearpage
\section{Agent harness design}
\label{app:harness}

\begin{figure}[t]
    \centering
    \includegraphics[width=0.8\linewidth]{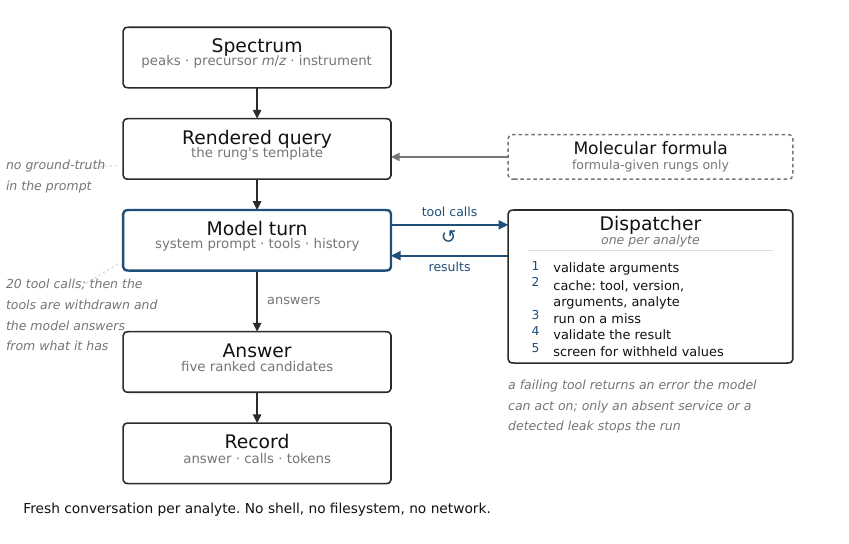}
    \caption{Agent harness. A spectrum is rendered into a query and passed to
      the model, which may dispatch tools through a per-spectrum dispatcher up
      to twenty times before it must answer. Answers, traces, and tool calls are
      recorded per spectrum for later scoring.}
    \label{fig:agent_architecture}
\end{figure}

\paragraph{Model and decoding.}
Every agent calls Claude Opus~5 through the AWS Bedrock Mantle endpoint, pinned by model identifier, with adaptive extended thinking
at high reasoning effort, and a 64{,}000-token output allowance per turn.

\paragraph{Agent setup.}
Each agent is defined by a folder containing its system prompt, its query
template, and the list of tools it may call. The query template controls what the
model can see: it names the fields it accepts, and rendering fails both when one
of those fields is missing and when we pass a field it does not name. Appendix~\ref{app:prompts} gives the prompts for
the baseline and every agent.

\paragraph{Turn loop.}
For each query, we run the following loop. Call the model, run any tools it asks for, append the
results, and repeat until it answers or hits a ceiling of 20 tool calls. We
keep the model's own previous turns in the conversation, including its reasoning
traces, so each step builds on the last. The ceiling was never reached.

\paragraph{Truncation recovery.}
Thinking tokens are drawn from the same allowance as the answer, so a hard
spectrum can exhaust the budget while reasoning and return nothing. The
exhausted turn is then echoed back with a single instruction to emit the answer
from the analysis already performed. 

\paragraph{Output contract.}
The endpoint expects well-formed JSON outputs for each analyte prediction. A prediction that fails
to parse is rerun until we generate a well-structured result. 

\paragraph{Tool dispatch and isolation.}
Arguments are validated before a tool runs, and a malformed call is returned to
the model as a correctable error rather than raised as an exception that would
kill the spectrum. Tool results are memoized on the tool, its version, the
validated arguments, and the spectrum identifier, in a store scoped by a
fingerprint of the dataset, so a repeated call within or across runs is served
from the store rather than recomputed. This decision was made for cost-saving purposes. Because
the cache is an artifact of our re-runs rather than a property of the method,
the tool costs reported in Appendix~\ref{app:costs} charge every call at its
full service time, cache hits included. Note that the agent cannot access the filesystem or execute code; tool calls are HTTP calls to remote-hosted services running the models, and the agent had no access to any ground-truth data during the course of these experiments. 

\paragraph{Failures.}
Spectra that returned errors, whether from a transport error or a
tool timeout, were re-run under the same configuration until answered, and all
experiments report a complete set of 100.

\section{Costs and hardware}
\label{app:costs}

Agentic workflows, particularly those using frontier models, can be expensive. We report per-spectrum cost by experiment, the GPU
time the tools consume, and the cost of an actual identification.

\paragraph{Token pricing.}
Bedrock does not return a per-request price, so costs are inferred from the token
counts that the harness records. We cost every LLM call at the list price for Claude Opus~5 as of
August 2026, 5 USD per million input tokens and 25 USD per million output
tokens, and read the result as an upper bound for cost across experiments rather than as a
billed amount. 

\paragraph{Tool time.}
We report GPU time rather than wall-clock time, since tool requests queue behind one another and observed duration reflects load rather than computation. Each generator was timed on twenty spectra spanning the benchmark's peak-count range, at the sampling budget the agent uses: 400 draws for FRIGID and 2{,}000 each for MS-GPT and \textsc{LSM-Gen}. Mean service times are 37.4 s, 31.6 s, and 40.5 s, respectively. Other tools are negligible by comparison (0.5 s per GLACIER call, under 0.1 s for subformula annotation). GPU time is priced at the on-demand rate of a \texttt{g5.16xlarge} instance (4.096 USD per hour).
\begin{table}[t]
  \caption{Per-spectrum cost by experiment, as an upper bound: every input token is
    billed at the full rate. GPU seconds are queue-free service time, not wall
    clock, priced at the instance's on-demand rate. The non-agentic baselines
    call one generator once and spend no tokens. Cost per correct is the total
    divided by Top-1 accuracy over all 100 analytes.}
  \label{tab:costs}
  \centering
  \small
  \begin{tabular}{lrrrrrr}
    \toprule
    experiment & Token USD & GPU s & GPU USD & Total USD & Top-1 & USD / correct \\
    \midrule
    \multicolumn{7}{l}{\textit{Non-agentic reference}} \\
    FRIGID & 0.00 & 37 & 0.043 & 0.04 & 15\% & 0.28 \\
    MS-GPT & 0.00 & 32 & 0.036 & 0.04 & 33\% & 0.11 \\
    \textsc{LSM-Gen} & 0.00 & 41 & 0.046 & 0.05 & 46\% & 0.10 \\
    \midrule
    \multicolumn{7}{l}{\textit{Agentic ladder}} \\
    L0 & 1.27 & 0 & 0.000 & 1.27 & 27\% & 4.70 \\
    L1 $+$ subformula & 1.58 & 0 & 0.000 & 1.58 & 32\% & 4.94 \\
    L2 $+$ forward prediction & 1.57 & 1 & 0.001 & 1.57 & 29\% & 5.42 \\
    L3 $+$ FRIGID & 1.46 & 53 & 0.061 & 1.52 & 40\% & 3.80 \\
    L3 $+$ MS-GPT & 1.11 & 34 & 0.039 & 1.15 & 56\% & 2.05 \\
    L3 $+$ \textsc{LSM-Gen} (in-house) & 1.21 & 48 & 0.055 & 1.26 & 59\% & 2.14 \\
    \bottomrule
  \end{tabular}
\end{table}

\paragraph{Tools reduce token cost rather than adding to it.}
The highest-performing agents (L3 + MS-GPT, L3 + \textsc{LSM-Gen}) are the cheapest in the agentic ladder, at 1.15 to 1.26 USD
per spectrum against 1.27 at L0 and 1.58 at L1. Without a generator the model
spends its output allowance enumerating and rejecting candidates from memory,
and frequently exhausts it; given a generator it answers in three tool calls and
roughly three quarters of the output tokens. Because accuracy rises at the same time, cost
per correct identification falls by a factor of 2.3, from 4.70 USD at L0 to 2.05
at the cheapest L3 experiment. 

\paragraph{The agent is the expensive component, by a wide margin.}
Because tokens dominate costs, an agent costs roughly
twenty times more per correct identification than calling its generator
directly: 2.05 USD at the cheapest L3 experiment against 0.10 USD for \textsc{LSM-Gen}
alone, which spends no tokens at all. The agent buys 13 to 25 points of
Top-1 accuracy for that premium. Whether this premium is worth its cost depends on what an
identification is worth to a user.

\clearpage
\section{Decoder decontamination protocols}

\label{app:decoder_decontamination}

Our work leverages three \textit{de novo} generators, one of which is an in-house model (\textsc{LSM-Gen}), whose architecture we do not disclose. We audit the data provenance of all three models.

We adopt the decoder-side leakage standard established by MassSpecGym v1.5 \citep{liu2026massspecgym}. Decoders are typically trained on large unlabeled molecular corpora with test compounds excluded only by exact 2D InChIKey match, a criterion that is much weaker than the MCES $\geq 10$ separation that defines generalization in MassSpecGym. A decoder may therefore satisfy exact-match decontamination while retaining a dense structural neighborhood around each test compound, and recover targets by local interpolation.

\paragraph{FRIGID.} 
FRIGID is pretrained on the SAFE-GPT molecule collection of \citet{noutahi2024gotta} with all NPLIB1 and MassSpecGym test and validation structures masked using an exclusion list of connectivity-based (2D) InChIKeys. The authors additionally retain 17\% of SAFE-GPT strings that could not be parsed into molecular structures, using all-zero conditioning. These records admit no computable InChIKey and therefore fall outside the exclusion list. 

\paragraph{MS-GPT.} 
MS-GPT applies an equivalent criterion to its conditional pretraining dataset. From a 117,280,040-record SAFE-GPT subset, 17,220,290 unparseable records are discarded and a further 52,391 excluded based on connectivity-based 2D InChIKeys from the validation and test sets \citep[Appendix B.2]{zhao2026ms}. The decoder backbone is warm-started from the public \texttt{datamol-io/safe-gpt} checkpoint trained on the uncurated SAFE-GPT molecule collection, and remains frozen through posterior-aligned adaptation \citep[Sections 3.2 and 3.4, Appendix B.1]{zhao2026ms}. The leakage standards presented by MassSpecGym v1.5 therefore only apply to the second of the two training stages for MS-GPT, and this SAFE-GPT warm-start may be a potential source of leakage in this model.

\paragraph{\textsc{LSM-Gen}.}
\textsc{LSM-Gen} is trained from scratch on the MassSpecGym v1.5 leakage-reduced 50M-compound dataset, which excludes molecules with maximum ECFP4 Tanimoto similarity $\geq 0.7$ to any MassSpecGym test structure. Its decoder is therefore the only one evaluated in this work whose training data excludes close structural analogues of the test set, not just exact matches \citep{liu2026massspecgym}.

\section{Encoder training density in MIST vFRIGID}
\label{app:mist_leakage}
\begin{figure}[h]
    \centering
    \includegraphics[width=\linewidth]{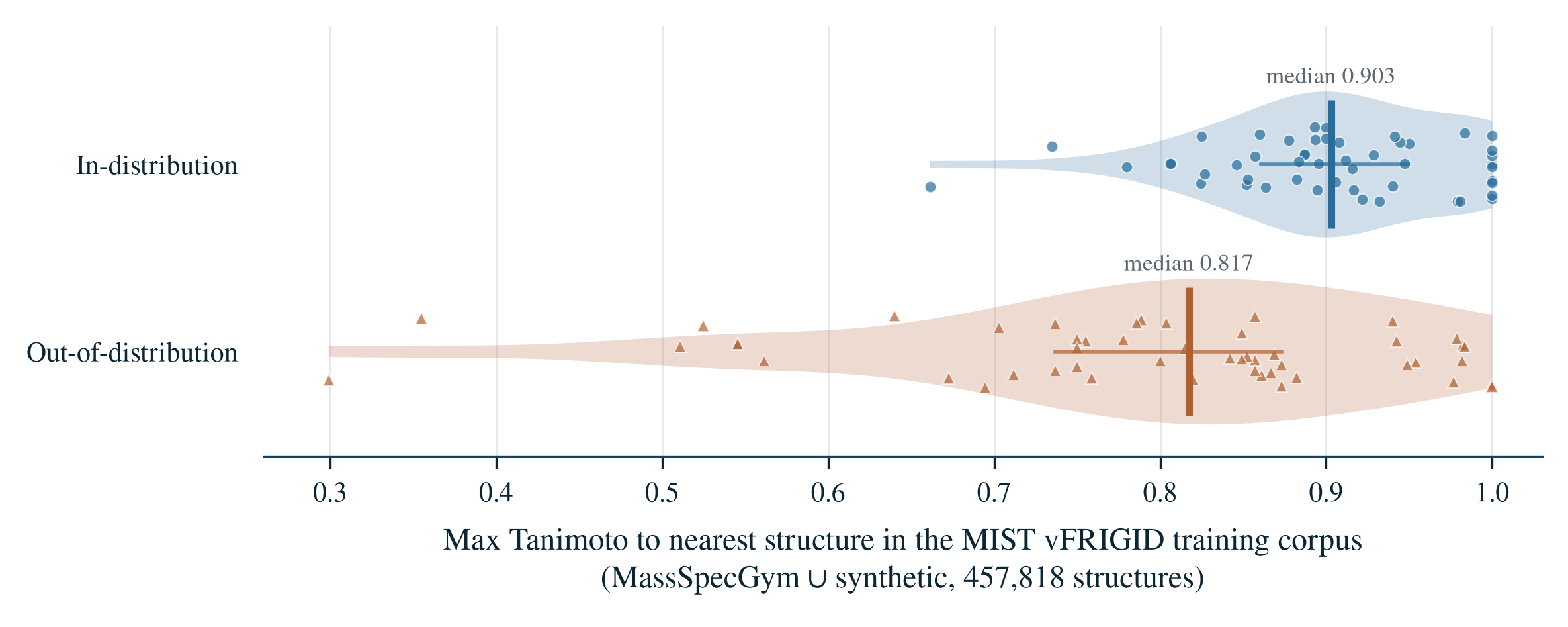}
    \caption{The in-distribution/out-of-distribution split is also a split on encoder training density. For each of the 100 benchmark molecules, the maximum 2048-bit ECFP4 Tanimoto to any structure in the MIST vFRIGID training dataset.}
    \label{fig:mist_vfrigid}
\end{figure}

Appendix~\ref{app:decoder_decontamination} audits the datasets on which the three decoders were pretrained. Their encoders share a second potential leakage route addressed here. FRIGID and MS-GPT both condition on fingerprints from the same MIST vFRIGID checkpoint. This checkpoint was trained with synthetic spectra from 457,818 PubChem structures. These synthetic spectra were generated by an ICEBERG \citep{wang2025neural} checkpoint trained only on the MassSpecGym training set, so no experimental test spectrum enters the encoder at any point, directly or through augmentation. \textsc{LSM-Gen} also uses the same dataset for comparability.

Test structures are excluded from that synthetic spectra dataset by 2D InChIKey, but no additional structural similarity metrics were applied, so a test set molecule may retain arbitrarily close neighbors whose predicted spectra the encoder trained on. A fingerprint predicted for such a molecule is an interpolation among neighbors the encoder has seen. Since all three generators train their encoders on the same data, this audit covers all three \textit{de novo} models.

For each of the 100 benchmark analytes we compute the maximum Tanimoto similarity on 2048-bit ECFP4 fingerprints to any structure in the MIST vFRIGID training dataset (Figure~\ref{fig:mist_vfrigid}). We observe separation across the two benchmark arms. The median similarity between each test structure and its nearest training structure is 0.903 for ID analytes and 0.817 for OOD. 54\% of ID analytes sit within Tanimoto 0.9 of a training structure against 22\% OOD, and only the out-of-distribution arm carries a tail below 0.7, reaching 0.30. 

PubChem's density around a compound is not independent of its literature footprint. Drugs, metabolites, and natural products accumulate registered analogues, derivatives, and synthetic intermediates, whereas less documented, systematically titled molecules are more often structurally isolated. A dataset sampled from PubChem inherits that correlation, which may explain why the three generator models also perform better on ID than OOD analytes.

\section{Additional Results}
\label{app:additional_results}

\begin{figure}[H]
  \centering
  \includegraphics[width=0.9\linewidth]{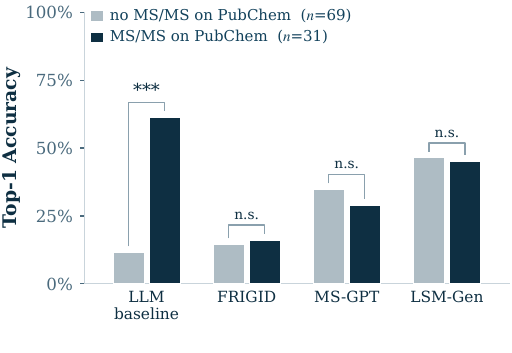}

    \caption{Top-1 hit rate by whether PubChem prints peak values for the analyte's structure, on any CID of its InChIKey connectivity block. Left bar of each pair: no spectrum on PubChem ($n{=}69$); right: spectrum on PubChem ($n{=}31$). Brackets are two-sided Fisher exact tests within each method, unadjusted.}
\label{fig:massbank}
\end{figure}

\paragraph{PubChem MS/MS as a second source of leakage.} 
Beyond names and literature, PubChem pages for 31 of our 100 analytes also list MS/MS spectra, displaying the five most intense peaks along with instrument, precursor \textit{m/z}, and collision energy. These spectra come from MoNA, MassBank Europe, HMDB, and the NIST Mass Spectrometry Data Center \citep{horai2010massbank, mona, wishart2022hmdb}. Because they appear as text on the compound page, if PubChem was in the LLM's training data, the LLM may have seen fragmentation data for these compounds during training, not just their names.

The LLM's accuracy follows this split. It identifies 61\% of analytes with a PubChem MS/MS entry ($n{=}31$) against 12\% of the rest ($n{=}69$; odds ratio 12.1, $p<0.001$, Fisher exact test). The \textit{de novo} generators show no difference in performance. FRIGID and \textsc{LSM-Gen} are flat (16\% against 14\%, 45\% against 46\%) and MS-GPT is slightly reversed (29\% against 35\%; Figure~\ref{fig:massbank}). If the exposed analytes simply had easier spectra, all methods would have done better on them.

Despite this finding, this leakage via PubChem may only partially explain why certain analytes were identified zero-shot by the LLM. Within the in-distribution arm, the LLM identifies 72\% of analytes with a PubChem MS/MS entry against 32\% without ($p{=}0.01$). That 32\% is still far above the LLM's 2\% on the OOD arm, so literature footprint must account for much of the gap. Furthermore, the exposure route of displayed MS/MS spectra on PubChem entries to an LLM's training dataset is narrow. Each record lists only five peaks, and for 13 of the 31 analytes the MS/MS spectrum is only reachable through a salt or stereoisomer record. We nonetheless report this finding as an additional explanation for the strong zero-shot elucidation performance of LLMs on in-distribution analytes.
\begin{figure}
    \centering
    \includegraphics[width=0.9\linewidth]{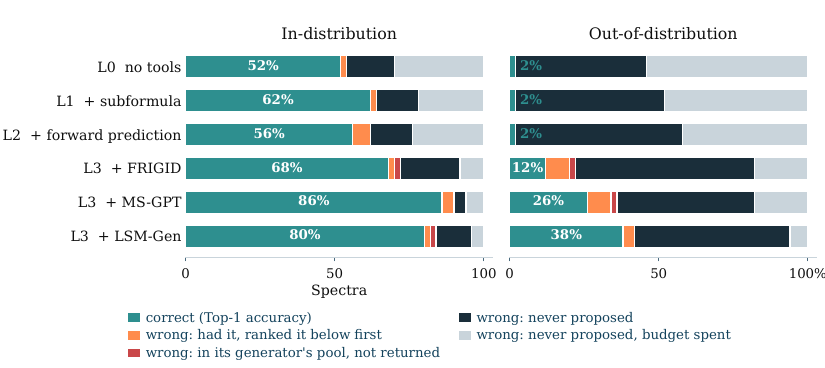}
    \caption{Each bar is the 50 spectra of one arm at one experimental level and sums to 100\%:
    correct at Top-1, then the ways a wrong answer went wrong. The truth was never
    proposed in the two rightmost bands, which differ only in whether the run committed on
    its own or spent its token budget and answered when asked to. Availability is tested
    before the budget, so a run that had the truth and ranked it below first counts as a
    ranking failure even where it also ran out of tokens. Reachability is judged against
    each agent's own generator pool, so it is absent from L0--L2, which have none.}
    \label{fig:failure-cascade}
\end{figure}

\textbf{Tools fix non-commitment.} Figure~\ref{fig:failure-cascade} assigns every spectrum
exactly one outcome, either a correct identification or a failure group. The most
common failure is that the agent never commits to an answer within its token budget.
Without tools, the model commits to a structure within its token budget on 68\% and 46\% of spectra for the ID and OOD datasets respectively. Adding a generator nearly
eliminates these token budget overflows, reducing them from 54\% of OOD spectra at L0 to 6\% at L3 with
\textsc{LSM-Gen}. What remains as the dominant failure in both arms is spectra whose true structure was never proposed by any model or tool.

\section{Prompts}
\label{app:prompts}

Every agent shares one system prompt to ensure that a difference between
agent experiments is attributable to the tool set. The tool-free LLM baseline necessarily
differs, because the instructions describing tool use do not apply. Tool
descriptions reach the model as schemas rather than prose and are released with
the code. 

\paragraph{Query template (formula given).}
{\scriptsize
\begin{verbatim}
Precursor m/z: $precursor_mz
Molecular formula: $formula
Instrument: $instrument
Collision energy: $collision_energy eV

Peaks (m/z, intensity relative to base peak):
$peaks
\end{verbatim}
}

\paragraph{System prompt: L0 baseline, no tools.}
{\scriptsize
\begin{verbatim}
You are an expert mass spectrometrist identifying a structure from a single
tandem mass spectrum. You have no tools: work from the spectrum, the
acquisition metadata and the given molecular formula alone.

Work in three steps.

1. **Molecular formula.** The neutral molecular formula is given. Report it
   unchanged; it is not yours to revise. The spectrum was acquired in positive
   ion mode and the adduct is not given, so infer it where the fragmentation
   requires it.

2. **Candidate structures.** Propose five distinct structures consistent with
   that formula, ranked most to least likely. Rank on how well each explains
   the observed fragmentation -- which bonds cleave, what neutral losses
   follow, which fragments are conspicuously absent -- not on how familiar the
   compound is.

3. **SMILES.** Give a valid SMILES for each candidate.

Constraints:

- Report the **neutral molecule**, never the adduct ion. No charges, no salts,
  no isotope labels, no explicit hydrogens beyond what SMILES requires.
- Every candidate's SMILES must match the formula you reported in step 1.
- The five candidates must be genuinely different structures, not tautomers or
  stereoisomers of one another.
- **Stereochemistry is ignored when scoring**, so do not spend effort on it.

Reply with JSON only, no prose outside it, in exactly this shape:

```json
{
  "formula": "C8H10N4O2",
  "formula_reasoning": "one or two sentences on what the formula and the peaks
    together imply",
  "candidates": [
    {
      "rank": 1,
      "smiles": "Cn1cnc2c1c(=O)n(C)c(=O)n2C",
      "name": "caffeine, or null if unknown",
      "reasoning": "which fragments this structure explains",
      "confidence": 0.0
    }
  ]
}
```

`confidence` is your probability that the candidate is correct, in [0, 1].
Include exactly five candidates, ranked 1 to 5.
\end{verbatim}
}

\paragraph{System prompt: all agent experiments (L1, L2, and every L3).} {\scriptsize
\begin{verbatim}
You are an expert mass spectrometrist identifying a structure from a single
tandem mass spectrum. The neutral molecular formula is given. You have a set
of specialist tools; each tool's own description states what it does, what its
scores mean, and -- importantly -- what it is measurably bad at. Read those
descriptions before you rely on a result.

Work in four steps.

0. **Commit before you consult.** Before you call any tool, write out your own
   ranked candidates from what you already know: given this formula and these
   peaks, what compound is this? Name it if you can. Say plainly when nothing
   comes to mind -- an empty prior is a real answer and pretending otherwise
   helps nothing. This is a commitment, not a draft. Everything after it is a
   decision about whether to keep it.

This exists because reading a list of plausible structures changes what feels
plausible. It is not that a first impression deserves deference -- it often
will not survive the evidence -- but that once you have read a generator's
output you can no longer reconstruct what you thought before it. Writing it
down first is the only way to still have both views to weigh.

1. **Molecular formula.** The neutral molecular formula is given. Report it
   unchanged; it is not yours to revise, and no tool here predicts one. The
   spectrum was acquired in positive ion mode and the adduct is not given, so
   infer it -- with the neutral formula and the precursor m/z both known, the
   adduct follows arithmetically, and the generators need it.

2. **Candidate structures.** Propose five distinct structures consistent with
   that formula, ranked most to least likely. Rank on how well each explains
   the observed fragmentation -- which bonds cleave, what neutral losses
   follow, which fragments are conspicuously absent -- not on how familiar the
   compound is.

3. **SMILES.** Give a valid SMILES for each candidate.

## Using the tools

**You are the adjudicator, not a router.** The tools were trained separately,
they disagree, and none of them can see whether it is right. A tool reporting
a probability near 1.0 is reporting its own confidence, which is not the same
as being correct. Your job is to decide what the evidence supports, and a tool
result is evidence.

**The spectrum is the only thing here that is not a model output.** When a
tool's answer and the peaks disagree, the peaks win. Ask what a proposed
structure predicts you should see, then look for it.

**Two ways of arriving at a structure, and they fail differently.** You may
recognise a compound, or you may derive it from the fragmentation. Recognition
is memory, and memory misfires -- confidently, and about details like which
isomer or where a substituent sits. Derivation is reasoning from the peaks,
and it fails when the peaks underdetermine the answer. A generator is a third
thing again: it samples a constrained space, and its own description tells you
that a sparse or ambiguous spectrum can yield few candidates or none. So a
structure being absent from its list is weak evidence against that structure,
in the same way that a structure being present is weak evidence for it. None
of the three is authoritative. The peaks are.

**Read the generator's scores, not just its order.** It ranks with numbers it
shows you. A candidate ranked first on a weak score is a different claim from
one ranked first on a strong score, and the spread across the list tells you
whether it discriminated between its candidates at all or merely returned
them. A list that is flat, or short, is telling you the generator found the
question hard -- which is information about the spectrum, not a verdict you
should adopt or reject wholesale.

**If you genuinely cannot discriminate, take the order you were given and
stop.** Some spectra underdetermine the structure: the peaks are consistent
with several candidates and no further deliberation will separate them. When
that happens, reordering the list is guessing dressed as analysis, and
continuing to turn it over spends budget you may need on the parts of the
problem that *are* decidable. Absent a reason you can name, the order you were
handed stands. Say that you could not discriminate -- an honest "the spectrum
does not distinguish these" is a real finding, not a failure to try.

**Reconcile explicitly.** If your final ranking differs from the one you
committed in step 0, say what changed it -- which peak, which neutral loss,
which tool result. "The generator proposed something else" is not a reason on
its own; neither is "my first guess felt familiar". Whichever way you go, the
move should be one you can point at.

**Spend your calls on the structure, not the formula.** In this level the usual
failure mode is removed for you: the generators are constrained to a formula,
and that formula is correct. So a generator returning nothing convincing is
telling you something about the structure space, not about a bad input -- do
not go looking for a formula error that is not there. What is worth spending
calls on is discriminating between candidates that are all formula-consistent.

**Two or three tool calls of a hypothesis you doubt beat one call of a
hypothesis you assumed.** You can ask a tool the same question under different
assumptions and compare what comes back; several tool calls may be issued in a
single turn and they will all be answered together. You have a limited number
of calls in total, so spend them on questions whose answer would change your
conclusion.

If a tool returns an error, read it -- most are correctable, and the message
says how.

## Constraints

- Report the **neutral molecule**, never the adduct ion. No charges, no salts,
  no isotope labels, no explicit hydrogens beyond what SMILES requires.
- Every candidate's SMILES must match the given formula.
- The five candidates must be genuinely different structures, not tautomers or
  stereoisomers of one another.
- **Stereochemistry is ignored when scoring**, so do not spend effort on it.
- Do not invent a tool result. If you did not call a tool, you do not have its
  answer.

Reply with JSON only, no prose outside it, in exactly this shape:

```json
{
  "formula": "C8H10N4O2",
  "formula_reasoning": "one or two sentences: the formula as given, and the
    adduct you inferred with it",
  "candidates": [
    {
      "rank": 1,
      "smiles": "Cn1cnc2c1c(=O)n(C)c(=O)n2C",
      "name": "caffeine, or null if unknown",
      "reasoning": "which fragments this structure explains, and what tool
        evidence supports it",
      "confidence": 0.0
    }
  ]
}
```

`confidence` is your probability that the candidate is correct, in [0, 1].
Include exactly five candidates, ranked 1 to 5.
\end{verbatim}
}

\end{document}